\documentclass[11pt,a4paper]{article}
\usepackage[hyperref]{conll-2019}
\usepackage{times}
\usepackage{amsmath,graphicx}
\usepackage{latexsym}
\usepackage{CJKutf8}
\usepackage[T1]{fontenc}
\usepackage{multirow}
\usepackage{amssymb}
\usepackage{subfig, graphicx}
\usepackage{url}

\definecolor{Orange}{rgb}{1.0, 0.43, 0}

\aclfinalcopy 

\title{Code-Switched Language Models \\Using Neural Based Synthetic Data from Parallel Sentences}

\author{Genta Indra Winata, Andrea Madotto, Chien-Sheng Wu, Pascale Fung \\
Center for Artificial Intelligence Research (CAiRE)\\
  Department of Electronic and Computer Engineering\\
  The Hong Kong University of Science and Technology, Clear Water Bay, Hong Kong\\
  \texttt{\{giwinata,amadotto,cwuak\}@connect.ust.hk}, \texttt{pascale@ece.ust.hk}}

\date{}

\begin{document}
\maketitle
\begin{abstract}
Training code-switched language models is difficult due to lack of data and complexity in the grammatical structure. Linguistic constraint theories have been used for decades to generate artificial code-switching sentences to cope with this issue. However, this require external word alignments or constituency parsers that create erroneous results on distant languages. We propose a sequence-to-sequence model using a copy mechanism to generate code-switching data by leveraging parallel monolingual translations from a limited source of code-switching data. The model learns how to combine words from parallel sentences and identifies when to switch one language to the other. Moreover, it captures code-switching constraints by attending and aligning the words in inputs, without requiring any external knowledge. Based on experimental results, the language model trained with the generated sentences achieves state-of-the-art performance and improves end-to-end automatic speech recognition.

\end{abstract}



\section{Introduction}
\textit{Code-switching} is a common linguistic phenomenon in multilingual communities, in which a person begins speaking or writing in one language and then switches to another in the same sentence.\footnote{Code-switching refers to mixing of languages following the definitions in~\citet{poplack1980sometimes}. We use ``intra-sentential code-switching" interchangeably with ``code-mixing".} 
It is motivated in response to social factors as a way of communicating in a multicultural society.
In its practice, code-switching varies due to the traditions, beliefs, and normative values in the respective communities. 
Linguists have studied the code-switching phenomenon and proposed a number of linguistic theories \cite{poplack1978syntactic,pfaff1979constraints,poplack1980sometimes, belazi1994code}. Code-switching is not produced indiscriminately, but follows syntactic constraints. Many linguists have formulated various constraints to define a general rule for code-switching \cite{poplack1978syntactic, poplack1980sometimes,belazi1994code}. However, these constraints cannot be postulated as a universal rule for all code-switching scenarios, especially for languages that are syntactically divergent \cite{berk1986linguistic}, such as English and Mandarin since they have word alignments with an inverted order. 

Building a language model (LM) and an automatic speech recognition (ASR) system that can handle intra-sentential code-switching is known to be a difficult research challenge.~The main reason lies in the unpredictability of code-switching points in an utterance and data scarcity.~Creating a large-scale code-switching dataset is also very expensive. Therefore, code-switching data generation methods to augment existing datasets are a useful workaround.


Existing methods that apply equivalence constraint theory to generate code-switching sentences \cite{li2012code, pratapa2018language} may suffer performance issues as they receive erroneous results from the word aligner and the part-of-speech (POS) tagger. Thus, this approach is not reliable and effective. Recently, \citet{garg2018code} proposed a SeqGAN-based model to generate code-switching sentences. Indeed, the model learns how to generate new synthetic sentences. However, the distribution of the generated sentences is very different from real code-switching data, which leads to underperforming results. 

To overcome the challenges in the existing works, we introduce a neural-based code-switching data generator model using pointer-generator networks (Pointer-Gen) \cite{SeeP17-1099} to learn code-switching constraints from a limited source of code-switching data and leverage their translations in both languages. Intuitively, the copy mechanism can be formulated as an end-to-end solution to copy words from parallel monolingual sentences by aligning and reordering the word positions to form a grammatical code-switching sentence. This method solves the two issues in the existing works by removing the dependence on the aligner or tagger, and generating new sentences with a similar distribution to the original dataset. Interestingly, this method can learn the alignment effectively without a word aligner or tagger. As an additional advantage, we demonstrate its interpretability by showing the attention weights learned by the model that represent the code-switching constraints.
Our contributions are summarized as follows:
\begin{itemize}
    \item We propose a language-agnostic method to generate code-switching sentences using a pointer-generator network \cite{SeeP17-1099} that learns when to switch and copy words from parallel sentences, without using external word alignments or constituency parsers. By using the generated data in the language model training, we achieve the state-of-the-art performance in perplexity and also improve the end-to-end ASR on an English-Mandarin code-switching dataset.
    \item We present an implementation applying the equivalence constraint theory to languages that have significantly different grammar structures, such as English and Mandarin, for sentence generation.~We also show the effectiveness of our neural-based approach in generating new code-switching sentences compared to the equivalence constraint and SeqGAN \cite{garg2018code}.
    \item We thoroughly analyze our generation results and further examine how our model identifies code-switching points to show its interpretability.
    
\end{itemize}


\section{Generating Code-Switching Data}
\label{sec:length}
In this section, we describe our proposed model to generate code-switching sentences using a pointer-generator network. Then, we briefly list the assumptions of the equivalence constraint (EC) theory, and explain our application of EC theory for sentence generation. We call the dominant language the matrix language ($L_1$) and the inserted language the embedded language ($L_2$), following the definitions from \citet{myers2001matrix}.~Let us define $Q = \{Q_1,..., Q_T\}$ as a set of $L_1$ sentences and $E = \{E_1,..., E_T\}$ as a set of $L_2$ sentences with $T$ number of sentences, where each $Q_t = \{q_{1,t},...,q_{m,t}\}$ and $E_t = \{e_{1,t},...,e_{n,t}\}$ are sentences with $m$ and $n$ words.~$E$ is the corresponding parallel sentences of $Q$. 



\subsection{Pointer-Gen}
Initially, Pointer-Gen was proposed to learn when to copy words directly from the input to the output in text summarization, and they have since been successfully applied to other natural language processing tasks, such as comment generation \cite{lin2019learning}. The Pointer-Gen leverages the information from the input to ensure high-quality generation, especially when the output sequence consists of elements from the input sequence, such as code-switching sequences.

\begin{figure*}[!t]
  \centering
  \includegraphics[width=0.87\linewidth]{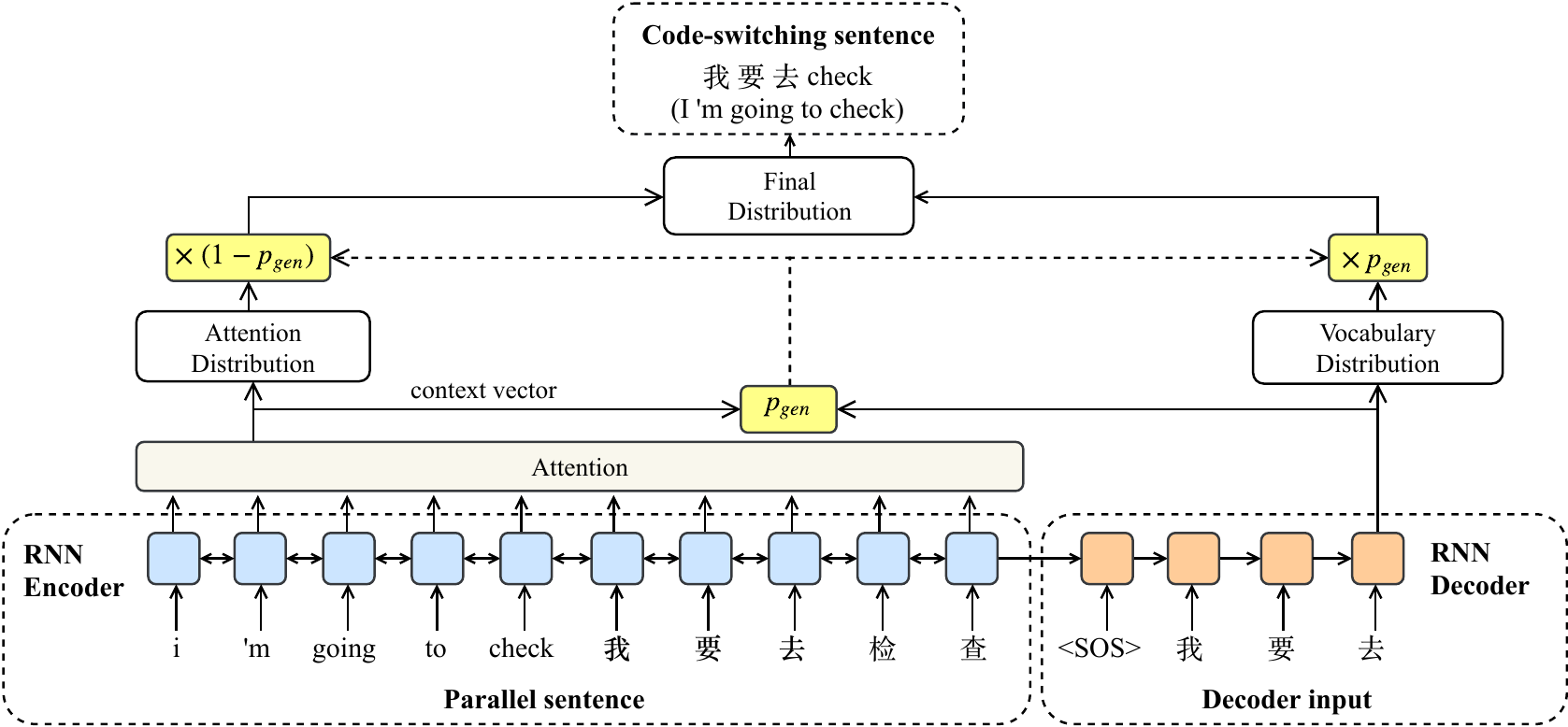}
  \caption{\textbf{Pointer-Gen} model, which includes an RNN encoder and RNN decoder. The parallel sentence is the input of the encoder, and in each decoding step, the decoder generates a new token.}
  \label{fig:pointer-generator}
\end{figure*}

We propose to use Pointer-Gen by leveraging parallel monolingual sentences to generate code-switching sentences. The approach is depicted in Figure \ref{fig:pointer-generator}. The pointer-generator model is trained from concatenated sequences of parallel sentences ($Q$,$E$) to generate code-switching sentences, constrained by code-switching texts. The words of the input are fed into the encoder. We use a bidirectional long short-term memory (LSTM), which, produces hidden state $h_t$ in each step $t$. The decoder is a unidirectional LSTM receiving the word embedding of the previous word. For each decoding step, a generation probability $p_{gen}$ $\in$ [0,1] is calculated, which weights the probability of generating words from the vocabulary, and copying words from the source text.~$p_{gen}$ is a soft gating probability to decide whether to generate the next token from the decoder or to copy the word from the input instead. The attention distribution $a_t$ is a standard attention with general scoring~\cite{luong2015effective}. It considers all encoder hidden states to derive the context vector. The vocabulary distribution $P_{voc}(w)$ is calculated by concatenating the decoder state $s_t$ and the context vector $h_t^*$:
\begin{equation}
p_{gen} = \sigma (w_{h^*}^T h_t^* + w_s^T s_t + w_x^T x_t + b_{ptr}),
\end{equation}
where $w_{h^*}, w_s, $ and $w_x$ are trainable parameters and $b_{ptr}$ is the scalar bias. The vocabulary distribution $P_{voc}(w)$ and the attention distribution $a^t$ are weighted and summed to obtain the final distribution $P(w)$, which is calculated as follows:
\begin{equation}
P(w) = p_{gen} P_{voc}(w) + (1 - p_{gen})\sum_{i:w_i=w}{a_i^t}.
\end{equation}
We use a beam search to select the $N$-best code-switching sentences.



 \begin{figure}[!t]
  \centering
  \includegraphics[width=\linewidth]{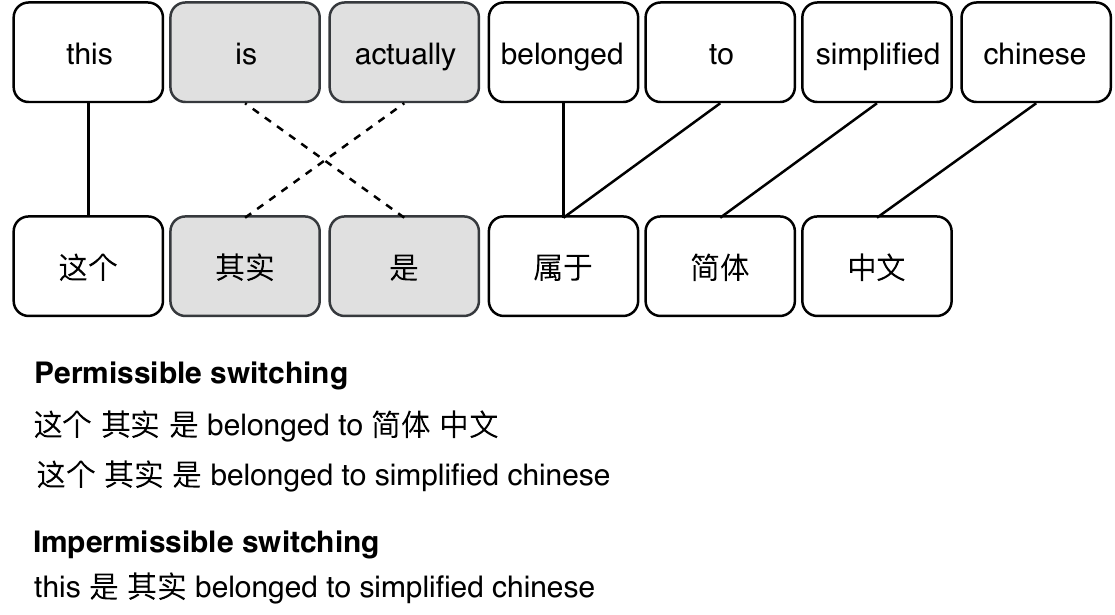}
  \caption{Example of equivalence constraint \cite{li2012code}. Solid lines show the alignment between the matrix language (top) and the embedded language (bottom).~The dotted lines denote impermissible switching.}
  \label{fig:eq-contraint}
\end{figure}

\subsection{Equivalence Constraint}
\begin{CJK*}{UTF8}{gbsn}
Studies on the EC \cite{poplack1980sometimes,poplack2013sometimes} show that code-switching only occurs where it does not violate the syntactic rules of either language. An example of a English-Mandarin mixed-language sentence generation is shown in Figure \ref{fig:eq-contraint}, where EC theory does not allow the word ``其实" to come after ``是" in Chinese, or the word ``is" to come after ``actually".  \citet{pratapa2018language} apply the EC in English-Spanish language modeling with a strong assumption. We are working with English and Mandarin, which have distinctive grammar structures (e.g., part-of-speech tags), so applying a constituency parser would give us erroneous results. Thus, we simplify sentences into a linear structure, and we allow lexical substitution on non-crossing alignments between parallel sentences. Alignments between an $L_1$ sentence $Q_t$ and an $L_2$ sentence $E_t$  comprise a source vector with indices~$u_t = \{a_1, a_2, ..., a_m\} \in \mathbb{W}^m$ that has a corresponding target vector $v_t = \{b_1, b_2, ..., b_m\} \in \mathbb{W}^m$, where $u$ is a sorted vector of indices in an ascending order. The alignment between $a_i$ and $b_i$ does not satisfy the constraint if there exists a pair of $a_j$ and $b_j$, where ($a_i < a_j$, and $b_i  > b_j$) or ($a_i > a_j$, and $b_i  < b_j$). If the switch occurs at this point, it changes the grammatical order in both languages; thus, this switch is not acceptable. During the generation step, we allow any switches that do not violate the constraint. We propose to generate synthetic code-switching data by the following steps:
\begin{enumerate}
\item Align the $L_1$ sentences $Q$ and $L_2$ sentences $E$ using \texttt{fast\_align}\footnote{The code implementation can be found at https://github.com/clab/fast\_align.}~ \cite{N13-1073}.~We use the mapping from the $L_1$ sentences to the $L_2$ sentences.
\item Permute alignments from step (1) and use them to generate new sequences by replacing the phrase in the $L_1$ sentence with the aligned phrase in the $L_2$ sentence.
\item Evaluate generated sequences from step (2) if they satisfy the EC theory.
\end{enumerate}
\end{CJK*}

\section{End-to-End Code-Switching ASR}
To show the effectiveness of our proposed method, we build a transformer-based end-to-end code-switching ASR system. 
The end-to-end ASR model accepts a spectrogram as the input, instead of log-Mel filterbank features~\cite{Zhou2018SyllableBasedSS}, and predicts characters. It consists of $N$ layers of an encoder and decoder. Convolutional layers are added to learn a universal audio representation and generate input embedding. We employ multi-head attention to allow the model to jointly attend to information from different representation subspaces at a different position.

For proficiency in recognizing individual languages, we train a multilingual ASR system trained from monolingual speech. The idea is to use it as a pretrained model and transfer the information while training the model with code-switching speech. This is an effective method to initialize the parameters of low-resource ASR such as code-switching. The catastrophic forgetting issue arises when we train one language after the other. Therefore, we solve the issue by applying a multi-task learning strategy. We jointly train speech from both languages by taking the same number of samples for each language in every batch to keep the information of both tasks.


In the inference time, we use beam search, selecting the best sub-sequence scored using the softmax probability of the characters. We define $P(Y)$ as the probability of the sentence. We incorporate language model probability $p_{lm}(Y)$ to select more natural code-switching sequences from generation candidates. A word count is added to avoid generating very short sentences. $P(Y)$ is calculated as follows:
\begin{equation}
P(Y) = \alpha P_{trans}(Y|X) + \beta p_{lm}(Y)  + \gamma \sqrt{wc(Y)}
\end{equation}
where $\alpha$ is the parameter to control the decoding probability from the probability of characters from the decoder $P_{trans}(Y|X)$, $\beta$ is the parameter to control the language model probability $p_{lm}(Y)$, and $\gamma$ is the parameter to control the effect of the word count $wc(Y)$.

\section{Experiments}
\subsection{Data Preparation}
We use speech data from SEAME Phase II, a conversational
English-Mandarin Chinese code-switching speech corpus that consists of spontaneously spoken interviews and conversations~\cite{SEAME2015}. We split the corpus following information from~\citet{winata2018code}. The details are depicted in  Table~\ref{data-statistics-phase-2}. We tokenize words using the Stanford NLP toolkit~\cite{manning-EtAl:2014:P14-5}. For monolingual speech datasets, we use HKUST~\cite{liu2006hkust}, comprising spontaneous Mandarin Chinese telephone speech recordings, and Common Voice, an open-accented English dataset collected by Mozilla.\footnote{The dataset is available at https://voice.mozilla.org/.}
We split Chinese words into characters to avoid word boundary issues, similarly to \citet{garg2018code}. We generate $L_1$ sentences and $L_2$ sentences by translating the training set of SEAME Phase II into English and Chinese using the Google NMT system (To enable reproduction of the results, we release the translated data).\footnote{We have attached the translated data in the Supplementary Materials.} Then, we use them to generate 270,531 new pieces of code-switching data, which is thrice the number of the training set. Table~\ref{generated-sequences} shows the statistics of the new generated sentences. To calculate the complexity of our real and generated code-switching corpora, we use the following measures:
\paragraph{Switch-Point Fraction (SPF)}~This measure calculates the number of switch-points in a sentence divided by the total number of word boundaries \cite{pratapa2018language}.~We define ``switch-point" as a point within the sentence at which the languages of words on either side are different.

\paragraph{Code Mixing Index (CMI)} This  measure counts the number of switches in a corpus \cite{gamback2014measuring}. At the utterance level, it can be computed by finding the most frequent language in the utterance and then counting the frequency of the words belonging to all other languages present. We compute this metric at the corpus level by averaging the values for all the sentences in a corpus.
The computation is shown as follows:
\begin{equation}
    C_u(x) = \frac{N(x) - max(\ell_i\in \ell\{t_{\ell_i}(x)\}) + P(x)}{N(x)},
\end{equation}
where $N(x)$ is the number of tokens of utterance $x$, $t_{\ell_i}$ is the tokens in language $\ell_i$, and $P(x)$ is the number of code-switching points in utterance $x$. We compute this metric at the corpus-level by averaging the values for all sentences.



\subsection{LM Training Strategy Comparison}
We generate code-switching sentences using three methods: EC theory, SeqGAN~\cite{garg2018code}, and Pointer-Gen. To find the best way of leveraging the generated data, we compare different training strategies as follows:
\begin{align*}
&\textbf{(1)} \text{ rCS}, \textbf{(2a)} \text{ EC},
\textbf{(2b)} \text{ SeqGAN},\\ &\textbf{(2c)} \text{ Pointer-Gen}, \textbf{ (3a)} \text{ EC \& rCS},\\ 
&\textbf{(3b)} \text{ SeqGAN \& rCS}, \textbf{ (3c)} \text{ Pointer-Gen \& rCS}\\
&\textbf{(4a)} \text{ EC} \rightarrow \text{rCS} \textbf{ (4b)} \text{ SeqGAN} \rightarrow \text{rCS},\\ &\textbf{(4c)} \text{ Pointer-Gen} \rightarrow \text{rCS}
\end{align*}
\textbf{(1)} is the baseline, training with real code-switching data.~\textbf{(2a--2c)} train with only augmented data. \textbf{(3a--3c)} train with the concatenation of augmented data with rCS.~\textbf{(4a--4c)} run a two-step training, first training the model only with augmented data and then fine-tuning with rCS. Our early hypothesis is that the results from \textbf{(2a)} and 
\textbf{(2b)} will not be as good as the baseline, but when we combine them, they will outperform the baseline. We expect the result of \textbf{(2c)} to be on par with \textbf{(1)}, since Pointer-Gen learns patterns from the rCS dataset, and generates sequences with similar code-switching points. 

\subsection{Experimental Setup}
In this section, we present the settings we use to generate code-switching data, and train our language model and end-to-end ASR.

\paragraph{Pointer-Gen} The pointer-generator model has 500-dimensional hidden states. We use 50k words as our vocabulary for the source and target. We optimize the training by Stochastic Gradient Descent with an initial learning rate of 1.0 and decay of 0.5. We generate the three best sequences using beam search with five beams, and sample 270,531 sentences, thrice the amount of the code-switched training data.

\paragraph{EC} We generate 270,531 sentences, thrice the amount of the code-switched training data. To make a fair comparison, we limit the number of switches to two for each sentence to get a similar number of code-switches (SPF and CMI) to Pointer-Gen.

\paragraph{SeqGAN} We implement the SeqGAN model using a PyTorch implementation\footnote{To implement SeqGAN, we use code from https://github.com/suragnair/seqGAN.}, and use our best trained LM baseline as the generator in SeqGAN.
We sample 270,531 sentences from the generator, thrice the amount of the code-switched training data (with a maximum sentence length of 20).

\begin{table}[!t]
\centering
\resizebox{0.4\textwidth}{!}{
\begin{tabular}{@{}rccc@{}}
\hline
\multicolumn{1}{|l|}{} & \multicolumn{1}{c|}{\textbf{Train}} & \multicolumn{1}{c|}{\textbf{Dev}} & \multicolumn{1}{c|}{\textbf{Test}} \\ \hline
\multicolumn{1}{|r|}{\# Speakers}               & \multicolumn{1}{c|}{138} & \multicolumn{1}{c|}{8} & \multicolumn{1}{c|}{8} \\
\multicolumn{1}{|r|}{\# Duration (hr)} & \multicolumn{1}{c|}{100.58} & \multicolumn{1}{c|}{5.56} & \multicolumn{1}{c|}{5.25} \\ 
\multicolumn{1}{|r|}{\# Utterances} & \multicolumn{1}{c|}{90,177} & \multicolumn{1}{c|}{5,722} & \multicolumn{1}{c|}{4,654} \\ 
\multicolumn{1}{|r|}{\# Tokens} & \multicolumn{1}{c|}{1.2M} & \multicolumn{1}{c|}{65K} & \multicolumn{1}{c|}{60K} \\ 
\multicolumn{1}{|r|}{CMI} & \multicolumn{1}{c|}{0.18}& \multicolumn{1}{c|}{0.22} & \multicolumn{1}{c|}{0.19} \\ 
\multicolumn{1}{|r|}{SPF} & \multicolumn{1}{c|}{0.15} & \multicolumn{1}{c|}{0.19} & \multicolumn{1}{c|}{0.17} \\ \hline
\end{tabular}
}
\caption{Data statistics of SEAME Phase II. The dataset is split by speaker ID.}
\label{data-statistics-phase-2}
\end{table}

\begin{table}[!t]
\centering
\resizebox{0.48\textwidth}{!}{
\begin{tabular}{|r|c|c|c|}
\hline
\multicolumn{1}{|l|}{} & \textbf{EC} & \textbf{SeqGAN} & \textbf{Pointer-Gen} \\ \hline
\# Utterances & 270,531 & 270,531 & 270,531 \\ 
\# Words & 3,040,202 & 2,981,078 & 2,922,941  \\ 
new unigram & 13.63\% & 34.67\% & 4.67\%\\ 
new bigram & 69.43\% & 80.33\% & 46.57\% \\
new trigram & 99.73\% & 141.56\% & 69.38\% \\ 
new four-gram & 121.04\% & 182.89\% & 85.07\% \\ 
CMI & 0.25 & 0.13 & 0.25 \\ 
SPF & 0.17 & 0.2 & 0.17 \\ \hline
\end{tabular}
}
\caption{Statistics of the generated data. The table shows the number of utterances and words, code-switches ratio, and percentage of new n-grams.}
\label{generated-sequences}
\end{table}

\begin{table*}[!t]
\centering
\resizebox{\textwidth}{!}{
\begin{tabular}{|l|c|c|c|c|c|c|c|c|c|c|}
\hline
\multirow{2}{*}{\textbf{Training Strategy}} & \multicolumn{2}{c|}{\textbf{Overall}} & \multicolumn{2}{c|}{\textbf{en-zh}} & \multicolumn{2}{c|}{\textbf{zh-en}} & \multicolumn{2}{c|}{\textbf{en-en}} & \multicolumn{2}{c|}{\textbf{zh-zh}}  \\ \cline{2-11} 
  & \textbf{valid} & \textbf{test} & \textbf{valid} & \textbf{test} & \textbf{valid} & \textbf{test} & \textbf{valid} & \textbf{test} & \textbf{valid} & \textbf{test} \\ \hline
\multicolumn{11}{|l|}{\textit{Only real code-switching data}} \\ \hline
\text{(1)} rCS & 72.89 & 65.71 & 7411.42 & 7857.75 & 120.41 & 130.21 & 29.31 & 29.61 & 244.88 & 246.71 \\ \hline
\multicolumn{11}{|l|}{\textit{Only generated data}} \\ \hline
\text{(2a)} EC & 115.98 & 96.54 & 32865.62 & 30580.89 & 107.22 & 109.10 & 28.24 & 28.2 & 1893.77 & 1971.68 \\ 
\text{(2b)} SeqGAN & 252.86 & 215.17 & 33719 & 37119.9 & 174.2 & 187.5 & 91.07 & 88 & 1799.74 & 1783.71 \\
\text{(2c)} Pointer-Gen & 72.78 & 64.67 & 7055.59 & 7473.68 & 119.56 & 133.39 & 27.77 & 27.67 & 234.16 & 235.34 \\ \hline
\multicolumn{11}{|l|}{\textit{Concatenate generated data with real code-switching data}} \\ \hline
\text{(3a)} EC \& rCS & 70.33 & 62.43 & 8955.79 & 9093.01 & 130.92 & 139.06 & 26.49 & 26.28 & 227.57 & 242.30 \\
\text{(3b)} SeqGAN \& rCS & 77.37 & 69.58 & 8477.44 & 9350.73 & 134.27 & 143.41 & 30.64 & 30.81 & 260.89 & 264.28 \\
\text{(3c)} Pointer-Gen \& rCS & 68.49 & 61.57 & 7146.08 & 7667.82 & 127.50 & 139.06 & 26.75 & 26.96 & 218.27 & 226.60\\ \hline
\multicolumn{11}{|l|}{\textit{Pretrain with generated data and fine-tune with real code-switching data}} \\ \hline
\text{(4a)} EC $\rightarrow$ rCS & 68.46 & 61.42 & 8200.78 & 8517.29 & 101.15 & 107.77 & 25.49 & 25.78 & 247.3 & 258.95 \\
\text{(4b)} SeqGAN $\rightarrow$ rCS & 70.61 & 64.03 & 6950.02 & 7694.2 & 114.82 & 122.84 & 28.5 & 28.73 & 236.94 & 244.62 \\
\textbf{(4c)} \textbf{Pointer-Gen $\rightarrow$ rCS} & \textbf{66.08} & \textbf{59.74} & \textbf{6620.76} & \textbf{7172.42} & \textbf{114.53} & \textbf{127.12} & \textbf{26.36} & \textbf{26.40} & \textbf{216.02} & \textbf{222.49} \\ \hline
\end{tabular}
}
\caption{Results of perplexity (PPL) on a valid set and test set for different training strategies. We report the overall PPL, and code-switching points \textbf{(en-zh)} and \textbf{(zh-en)}, as well as the monolingual segments PPL  \textbf{(en-en)} and  \textbf{(zh-zh)}.}
\label{lm-results}
\end{table*}

\begin{CJK*}{UTF8}{gbsn}
\begin{figure*}[!t]
  \centering  \includegraphics[width=\linewidth]{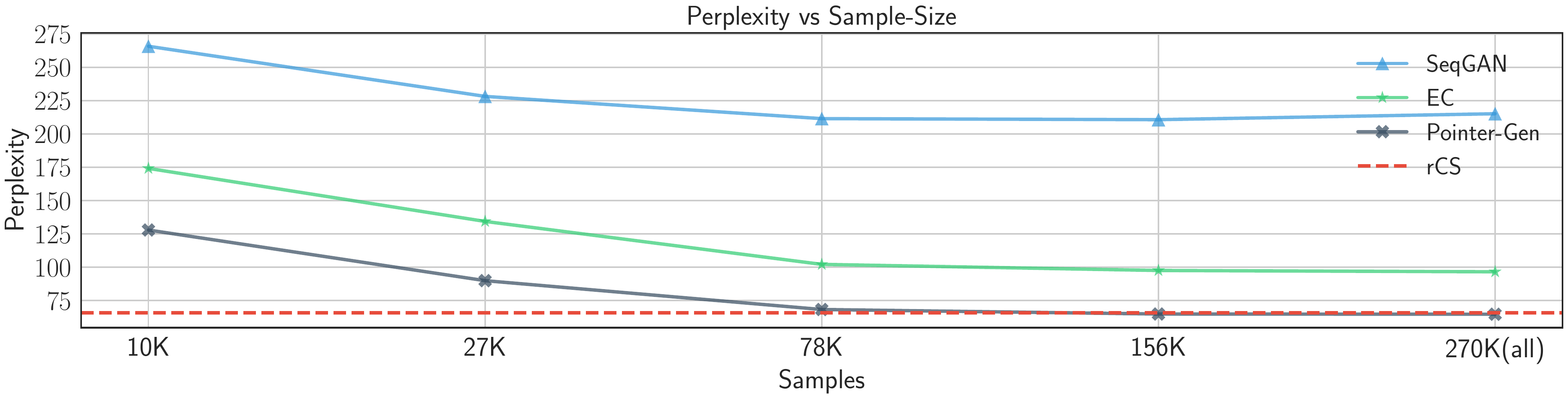}
  \caption{Results of perplexity (PPL) on different numbers of generated samples. The graph shows that Pointer-Gen attains a close performance to the real training data, and outperforms \textit{SeqGAN} and \textit{EC}.}
  \label{fig:samples}
\end{figure*}
\end{CJK*}

\paragraph{LM} In this work, we focus on sentence generation, so we evaluate our data with the same two-layer LSTM LM for comparison. It is trained using a two-layer LSTM with a hidden size of 200 and unrolled for 35 steps. The embedding size is equal to the LSTM hidden size for weight tying \cite{press2017using}. We optimize our model using SGD with an initial learning rate of 20. If there is no improvement during the evaluation, we reduce the learning rate by a factor of 0.75. In each step, we apply a dropout to both the embedding layer and recurrent network. The gradient is clipped to a maximum of 0.25. We optimize the validation loss and apply an early stopping procedure after five iterations without any improvements. In the fine-tuning step of training strategies \textbf{(4a--4c)}, the initial learning rate is set to 1.

\begin{CJK*}{UTF8}{gbsn}
\paragraph{End-to-end ASR} We convert the inputs into normalized frame-wise spectrograms from 16-kHz audio. Our transformer model consists of two encoder and decoder layers.~An Adam optimizer and Noam warmup are used for training with an initial learning rate of 1e-4. The model has a hidden size of 1024, a key dimension of 64, and a value dimension of 64. The training data are randomly shuffled every epoch. Our character set is the concatenation of English letters, Chinese characters found in the corpus, spaces, and apostrophes.
In the multilingual ASR pretraining, we train the model for 18 epochs. Since the sizes of the datasets are different, we over-sample the smaller dataset. The fine-tuning step takes place after the pretraining using code-switching data.~In the inference time, we explore the hypothesis using beam search with eight beams and a batch size of 1.
\end{CJK*}

\begin{CJK*}{UTF8}{gbsn}
\begin{figure*}[!t]
  \centering
  \includegraphics[width=\linewidth]{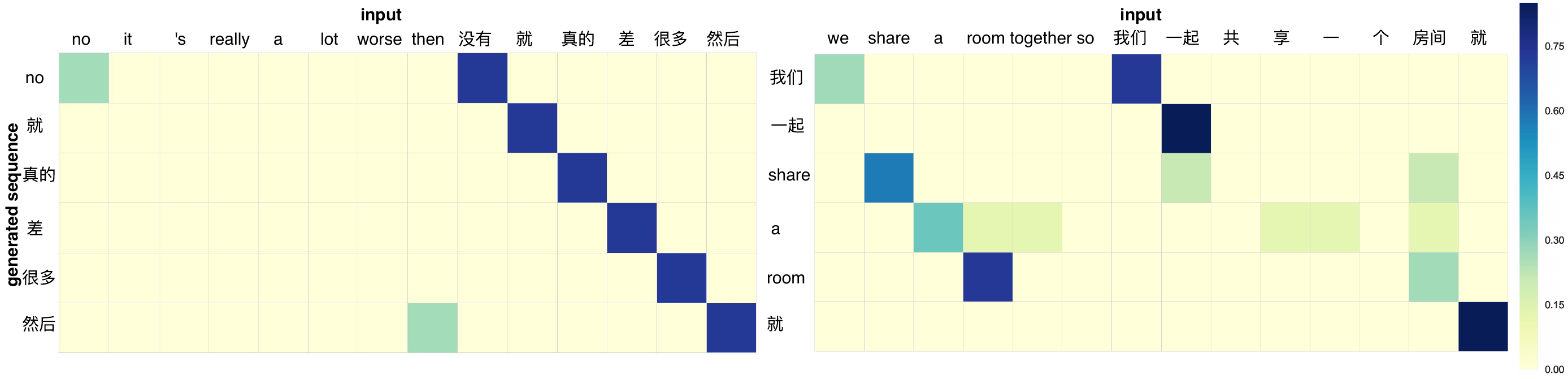}
  \caption{The visualization of pointer-generator attention weights on input words in each time-step during the inference time. The y-axis indicates the generated sequence, and the x-axis indicates the word input. In this figure, we show the code-switching points when our model attends to words in the $L_1$ and $L_2$ sentences: \textbf{left:} (``no",``没有") and (``then",``然后"), \textbf{right:} (``we",``我们"), (``share", ``一起") and (``room",``房间").}
  \label{fig:attention}
\end{figure*}
\end{CJK*}

\subsection{Evaluation Metrics}
We employ the following metrics to measure the performance of our models.

\paragraph{Token-level Perplexity (PPL)}
For the LM, we calculate the PPL of characters in Mandarin Chinese and words in English. The reason is that some Chinese words inside the SEAME corpus are not well tokenized, and tokenization results are not consistent. Using characters instead of words in Chinese can alleviate word boundary issues. The PPL is calculated by taking the exponential of the sum of losses. To show the effectiveness of our approach in calculating the probability of the switching, we split the perplexity computation into monolingual segments \textbf{(en-en)} and \textbf{(zh-zh)}, and code-switching segments \textbf{(en-zh)} and \textbf{(zh-en)}.

\paragraph{Character Error Rate (CER)} For our ASR, we compute the overall CER and also show the individual CERs for Mandarin Chinese \textbf{(zh)} and English \textbf{(en)}. The metric calculates the distance of two sequences as the \textit{Levenshtein Distance}. 


\section{Results \& Discussion}
\begin{CJK*}{UTF8}{gbsn}
\paragraph{LM} In Table \ref{lm-results}, we can see the perplexities of the test set evaluated on different training strategies. Pointer-Gen consistently performs better than state-of-the-art models such as \textit{EC} and \textit{SeqGAN}. Comparing the results of models trained using only generated samples, \textbf{(2a-2b)} leads to the undesirable results that are also mentioned by~\citet{pratapa2018language}, but it does not apply to Pointer-Gen (2c). We can achieve a similar results with the model trained using only real code-switching data, rCS. This demonstrates the quality of our data generated using Pointer-Gen. In general, combining any generated samples with real code-switching data improves the language model performance for both code-switching segments and monolingual segments. Applying concatenation is less effective than the two-step training strategy. Moreover, applying the two-step training strategy achieves the state-of-the-art performance. 

As shown in Table \ref{generated-sequences}, we generate new n-grams including code-switching phrases. This leads us to a more robust model, trained with both generated data and real code-switching data. We can see clearly that Pointer-Gen-generated samples have a distribution more similar to the real code-switching data compared with \textit{SeqGAN}, which shows the advantage of our proposed method.

\paragraph{Effect of Data Size}
To understand the importance of data size, we train our model with different amounts of generated data. Figure \ref{fig:samples} shows the PPL of the models with different amounts of generated data. An interesting finding is that our model trained with only 78K samples of Pointer-Gen data (same number of samples as rCS) achieves a similar PPL to the model trained with only rCS, while SeqGAN and EC have a significantly higher PPL. We can also see that 10K samples of Pointer-Gen data is as good as 270K samples of EC data. In general, the number of samples is positively correlated with the improvement in performance.


\begin{table}[!t]
\centering
\resizebox{0.49\textwidth}{!}{
\begin{tabular}{|l|c|c|c|}
\hline
\textbf{Model} & \textbf{Overall} & \textbf{en} & \textbf{zh}\\ \hline
Baseline & 34.40\% & 41.79\% & 35.94\% \\
+ Pre-training & 32.76\% & 40.06\% & 32.44\% \\
\hspace{3mm}\text{+ LM (rCS)} & 32.25\% & 39.45\% & 31.90\% \\ \hline
\hspace{3mm}\textbf{+ LM (Pointer-Gen $\rightarrow$ rCS)} & \textbf{31.07\%} & \textbf{38.39\%} & \textbf{30.85\%} \\ \hline
\end{tabular}
}
\caption{ASR evaluation, showing the performance on all sequences (Overall), English segments (en), and Mandarin Chinese segments (zh).}
\label{asr-evaluation}
\end{table}

\paragraph{ASR Evaluation}
We evaluate our proposed sentence generation method on an end-to-end ASR system. Table \ref{asr-evaluation} shows the CER of our ASR systems, as well as the individual CER on each language. Based on the experimental results, pretraining is able to reduce the error rate by 1.64\%, as it corrects the spelling mistakes in the prediction.~After we add LM (rCS) to the decoding step, the error rate can be reduced to 32.25\%. Finally, we replace the LM with LM (Pointer-Gen $\rightarrow$ rCS), and it further decreases the error rate by 1.18\%.

\begin{CJK*}{UTF8}{gbsn}
\begin{table*}[!t]
\centering
\resizebox{0.91\textwidth}{!}{
\begin{tabular}{|c|c|c|c|l|}
\hline
\multicolumn{2}{|c|}{\textbf{rCS}} & \multicolumn{3}{c|}{\textbf{Pointer-Gen}} \\ \hline
\textbf{POS tags} & \textbf{ratio} & \textbf{POS tags} & \multicolumn{1}{c|}{\textbf{ratio}} & \multicolumn{1}{l|}{\textbf{examples}} \\ \hline
\multicolumn{5}{|c|}{\textbf{English}} \\ \hline
\begin{tabular}[c]{@{}c@{}}NN \\ (noun)\end{tabular} & 56.16\% & \begin{tabular}[c]{@{}c@{}}NN \\ (noun)\end{tabular} & \multicolumn{1}{r|}{55.45\%} & \begin{tabular}[c]{@{}c@{}}\text{那} \text{个} \textbf{\textcolor{Orange}{consumer}} \text{是} \text{不} \\ (that consumer is not)\end{tabular} \\ \hline
\begin{tabular}[c]{@{}c@{}}RB \\ (adverb)\end{tabular} & 10.34\% & \begin{tabular}[c]{@{}c@{}}RB \\ (adverb)\end{tabular} & \multicolumn{1}{r|}{10.14\%} & \begin{tabular}[c]{@{}l@{}}okay \textbf{\textcolor{Orange}{so}} \text{其} \text{实} \\ (okay so its real)\end{tabular} \\ \hline
\begin{tabular}[c]{@{}c@{}}JJ \\ (adjective)\end{tabular} & 7.04\% & \begin{tabular}[c]{@{}c@{}}JJ \\ (adjective)\end{tabular} & \multicolumn{1}{r|}{7.16\%} & \begin{tabular}[c]{@{}l@{}}\text{我} \text{很} \textbf{\textcolor{Orange}{jealous}} \text{的} \text{每} \text{次} \\ (i am very jealous every time)\end{tabular} \\ \hline
\begin{tabular}[c]{@{}c@{}}VB \\ (verb)\end{tabular} & 5.88\% & \begin{tabular}[c]{@{}c@{}}VB \\ (verb)\end{tabular} & \multicolumn{1}{r|}{5.89\%} & \begin{tabular}[c]{@{}l@{}}\textbf{\textcolor{Orange}{compared}} \text{这} \text{个} \\ (compared to this)\end{tabular} \\ \hline \hline
\multicolumn{5}{|c|}{\textbf{Chinese}} \\ \hline
\begin{tabular}[c]{@{}c@{}}VV \\ (other verbs)\end{tabular} & 23.77\% & \begin{tabular}[c]{@{}c@{}}VV \\ (other verbs)\end{tabular} & 23.72\% & \begin{tabular}[c]{@{}l@{}}\text{讲} \text{的} \text{要} \textbf{\textcolor{Orange}{用}} microsoft word \\ (i want to use microsoft word)\end{tabular} \\ \hline
\begin{tabular}[c]{@{}c@{}}M \\ (measure word)\end{tabular} & 16.83\% & \begin{tabular}[c]{@{}c@{}}M \\ (measure word)\end{tabular} & 16.49\% & \begin{tabular}[c]{@{}l@{}}\text{我} \text{们} \text{有} \text{这} \textbf{\textcolor{Orange}{个}} god of war \\ (we have this god of war)\end{tabular} \\ \hline
\begin{tabular}[c]{@{}c@{}}DEG \\ (associative)\end{tabular} & 9.12\% & \begin{tabular}[c]{@{}c@{}}DEG \\ (associative)\end{tabular} & 9.13\% & \begin{tabular}[c]{@{}l@{}}\text{我} \text{们} \textbf{\textcolor{Orange}{的}} result \\ (our result)\end{tabular} \\ \hline
\begin{tabular}[c]{@{}c@{}}NN \\ (common noun)\end{tabular} & 9.08\% & \begin{tabular}[c]{@{}c@{}}NN \\ (common noun)\end{tabular} & 8.93\% & \begin{tabular}[c]{@{}l@{}}\text{我} \text{应} \text{该} \text{不} \text{会} \text{讲} \textbf{\textcolor{Orange}{话}} because intimidated by another\\ (i shouldn’t talk because intimidated by another)\end{tabular} \\ \hline
\end{tabular}
}
\caption{The most common English and Mandarin Chinese part-of-speech tags that trigger code-switching. We report the frequency ratio from \textbf{Pointer-Gen}-generated sentences compared to the real code-switching data. We also provide an example for each POS tag.}
\label{data-statistics-word-trigger}
\end{table*}
\end{CJK*}

\paragraph{Model Interpretability}
\begin{CJK*}{UTF8}{gbsn}
We can interpret a Pointer-Gen model by extracting its attention matrices and then analyzing the activation scores. We show the visualization of the attention weights in Figure \ref{fig:attention}.~The square in the heatmap corresponds to the attention score of an input word. In each time-step, the attention scores are used to select words to be generated. As we can observe in the figure, in some cases, our model attends to words that are translations of each other, for example, the words (``no",``没有"), (``then",``然后") , (``we",``我们"), (``share", ``一起"), and (``room",``房间"). This indicates the model can identify code-switching points, word alignments, and translations without being given any explicit information.

\paragraph{Code-Switching Patterns}
Table \ref{data-statistics-word-trigger} shows the most common English and Mandarin Chinese POS tags that trigger code-switching.~The distribution of word triggers in the Pointer-Gen data are similar to the real code-switching data, indicating our model's ability to learn similar code-switching points. Nouns are the most frequent English word triggers. They are used to construct an optimal interaction by using cognate words and to avoid confusion.~Also, English adverbs such as ``then" and ``so" are phrase or sentence connectors between two language phrases for intra-sentential and inter-sentential code-switching. On the other hand, Chinese transitional words such as the measure word ``个" or associative word ``的" are frequently used as inter-lingual word associations.




\end{CJK*}


\end{CJK*}



\section{Related Work}
\label{sec:related-work}
Code-switching language modeling research has been focused on building a model that handles mixed-language sentences and on generating synthetic data to solve the data scarcity issue. The first statistical approach using a linguistic theory was introduced by~\citet{li2012code}, who adapted the EC on monolingual sentence pairs during the decoding step of an ASR system. ~\citet{ying2014language} implemented a functional-head constraint lattice parser with a weighted finite-state transducer to reduce the search space on a code-switching ASR system. Then, \citet{adel2013recurrent} extended recurrent neural networks (RNNs) by adding POS information to the input layer and a factorized output layer with a language identifier. The factorized RNNs were also combined with an n-gram backoff model using linear interpolation \cite{adel2013combination}, and syntactic and semantic features were added to them~\cite{adel2015syntactic}.~\citet{baheti2017curriculum} adapted an effective curriculum learning by training a network with monolingual corpora of two languages, and subsequently trained on code-switched data.~A further investigation of EC and curriculum learning showed an improvement in English-Spanish language modeling \cite{pratapa2018language}, and a multi-task learning approach was introduced to train the syntax representation of languages by constraining the language generator \cite{winata2018code}. \citet{garg2018code} proposed to use SeqGAN \cite{yu2017seqgan} for generating new mixed-language sequences. \citet{winata2018bilingual} leveraged character representations to address out-of-vocabulary words in the code-switching named entity recognition. Finally, \citet{winata2019learning} proposed a method to represent code-switching sentence using language-agnostic meta-representations.


\section{Conclusion}
\label{sec:conclusion}
We propose a novel method for generating synthetic code-switching sentences using Pointer-Gen by learning how to copy words from parallel corpora. Our model can learn code-switching points by attending to input words and aligning the parallel words, without requiring any word alignments or constituency parsers. More importantly, it can be effectively used for languages that are syntactically different, such as English and Mandarin Chinese. Our language model trained using  outperforms equivalence constraint theory-based models.~We also show that the learned language model can be used to improve the performance of an end-to-end automatic speech recognition system.
  
\section*{Acknowledgments}
This work has been partially funded by ITF/319/16FP and MRP/055/18 of the Innovation Technology Commission, the Hong Kong SAR Government, and School of Engineering Ph.D. Fellowship Award, the Hong Kong University of Science and Technology, and RDC 1718050-0 of EMOS.AI. We sincerely thank the three anonymous reviewers for their insightful comments on our paper.
  
\bibliography{conll-2019}
\bibliographystyle{acl_natbib}

\end{document}